# The control architecture of a spherical robot for Minimally Invasive Surgery

Gabriela Rus[1][0000-0002-1751-8952], Nadim Al Hajjar[3][0000-0001-5986-1233], Paul Tucan[1][0000-0001-5660-8259], Ionut Zima[1][0009-0007-0483-7691], Calin Vaida[1][0000-0003-2822-9790], Corina Radu[4][0000-0003-0005-0262], Daniela Jucan[1][0009-0004-0219-9858] Damien Chablat[1,5][0000-0001-7847-6162], Tiberiu Antal[1][0000-0002-0042-5258], Doina Pisla[1,2][0000-0001-7014-9431]*

[1] CESTER, Research Center for Industrial Robots Simulation and Testing, Technical University of Cluj-Napoca, 400641 Cluj-Napoca, Romania
[2] Technical Sciences Academy of Romania, 26 Dacia Blvd, 030167 Bucharest, Romania
[3] Department of Surgery, "Iuliu Hatieganu" University of Medicine and Pharmacy, 400347 Cluj-Napoca, Romania
[4] Department of Internal Medicine, "Iuliu Hatieganu" University of Medicine and Pharmacy, 400347 Cluj-Napoca, Romania
[5] École Centrale Nantes, Nantes Université, CNRS, LS2N, UMR 6004, F-44000 Nantes, France
*Corresponding author

**Abstract** Control systems used in Minimally Invasive Surgery (MIS) play a crucial role in ensuring precision and safety throughout procedures. This paper presents a control architecture developed for a robotic system designed for MIS operations. The modular structure of the control system allows for compatibility with a range of procedures in abdominal and thoracic regions. The proposed control system, employing the master-slave concept, is presented alongside the experimental model. Functional validation is obtained by performing a Siemens NX simulation and comparing the results with several experimental runs using the experimental model of the robot. With its compact size and stiffness, the system holds promise for integration with other robotic systems. Future efforts will be dedicated to exploring and optimizing this potential collaboration to enhance the overall capabilities of robotic-assisted surgery.

## 1 Introduction

Due to its advantages such as: lower risk infection, faster recovery, and better outcomes, MIS poses the potential to become the future for an increasing number of surgeries, reason for which many innovative solutions like specialized instruments, platforms and access devices, imaging technologies and robotic assistance, have been developed [1]. Particular attention received the robotic assisted MIS, which enhances the interventions, providing advantages like increased precision and dexterity, improved agronomy and reduced fatigue for surgeons [2] [3] [4]. Robotic assisted MIS prioritizes patient well-being by reducing trauma, pain, and recovery time, aligning with patient-centered care principles [5][6].

The integration of robots in surgery began at the end of the 20th century with the da Vinci Surgical System. The system has revolutionized minimally invasive surgeries, offer-ing 360-degree Endo wrist, enhanced dexterity, and ergonomic design [7]. Other systems include Senhance Surgical Robotic System composed of three or four manipulative arms and a surgeon console equipped with eye tracking camera [8] and Versius composed of open surgeon console with hand controllers to manipulate the arms and modular robotics arms [9].

A critical component of these structures is represented by the control architecture which assures the functionality, accuracy and safety for the entire structure.

For example, the authors of [10] presents a surgical robot that can perform remote vascular surgery using a fuzzy PID control strategy to ensure the accuracy and stability of the master-slave operation. In [11] is proposed a master-slave robotically controlled telemanipulation. Benchtop tests were conducted to evaluate the accuracy of robotic instrument movements, the quality of force feedback, and the overall performance of the system. [12] discussed the design and implementation of a control system for a novel parallel robot specifically used in prostate biopsy.

A parallel structure with 3 DOF, which could form the basis for robots used in minimally invasive surgical operations, in terms of workspace and singularities, is presented in [13].

This paper presents the control architecture of a spherical robot designed for Minimally Invasive Surgery (MIS). The control and command structure are designed to be adaptable and modular, with the capability to cater to different MIS procedures, including those targeting the abdomen (e.g. for pancreatic and liver procedures), as well as thoracic procedures (e.g. for the esophagus). The objective is to validate the functionality of the system by conducting simulations in Siemens NX and comparing the outcomes with experimental trials using the experimental model of the robot.

Following the introduction section, the paper is structured as follows: Section 2 outlines the proposed robotic structure, Section 3 describes the control architecture, and Section 4 presents the functional validation. Section 5 is dedicated to discussion and conclusions.

## 2      Materials and methods

### 2.1    Kinematic scheme of the proposed system

The RCM is a fixed point around which a surgical instrument or robotic device pivots during minimally invasive procedures, enabling precise and stable movements within the body. Achieving the RCM in robotic systems for MIS can be obtained through various methods [14]. However, the chosen approach for the proposed system involves utilizing a robot architecture that kinematically constrains the entry point into the body.
This approach not only ensures precise alignment between the robot and the patient before instrument insertion but also maintains safe operation throughout the surgery, all without impacting the control system (it doesn't need additional actuators to maintain the RCM in a fixed point). This method ensures precise alignment before instrument insertion and maintains safe behavior throughout the surgery.

As depicted in Fig. 1, the robotic system features a spherical robot with 3DOF and RCM in the center of the sphere ($X_{RCM}$, $Y_{RCM}$, $Z_{RCM}$) defined by a circular guide of radius

R, allowing for the attachment of either active instruments or an endoscopic camera [15]. This system is controlled by one revolute joint M1 ($q_1$) and two prismatic joints M2 and M3 ($q_2$ and $q_3$). These movements enable precise positioning of the instrument under the insertion point (trocar), facilitating accurate and minimally invasive surgical procedures. Through the rotational movement executed by $q_1$ and the circular prismatic movement executed by $q_2$, the instrument is positioned under the insertion point (trocar), while the additional action of $q_3$ enables the insertion of the instrument (end effector – E (X, Y, Z)) through the insertion point.

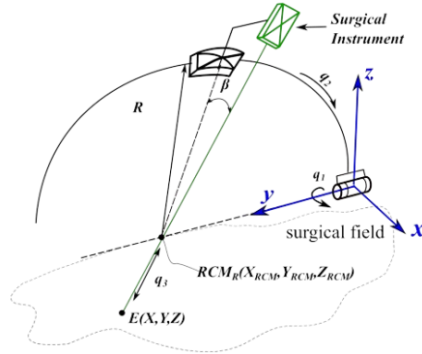

**Fig.1** Kinematic scheme of the spherical robot

The final equations for the kinematical model are [16]

- Inverse geometric model

$$\begin{cases} q_1 = \theta \\ q_2 = R\left(\psi + \frac{\pi}{2}\right) \\ q_3 = (R - m) + ins \end{cases} \quad (1)$$

- Forward geometric model

$$\begin{cases} X = X_{RCM} - (q_3 - R + m) \cdot \cos\left(\frac{q_2}{R} - \frac{\pi}{2}\right) \cdot \sin(q_1) \\ Y = Y_{RCM} - (q_3 - R + m) \cdot \sin\left(\frac{q_2}{R} - \frac{\pi}{2}\right) \\ Z = Z_{RCM} - (q_3 - R + m) \cdot \cos\left(\frac{q_2}{R} - \frac{\pi}{2}\right) \cdot \cos(q_1) \end{cases} \quad (2)$$

### 2.2 Experimental model

The final design of the spherical robot includes the development of a structure on the vertical axis, with the size of the robot being influenced by the dimensions of the chosen instrument. Considering that active instruments usually have a length between 500-600 mm and the endoscopic camera between 400-500 mm, the spherical rail should have a radius of 300 mm. The system is designed to allow interchangeability between instruments and endoscopic camera. As a result of the spherical structure and dimensional constraints, a compact and easily integrable robotic structure in the operating room has been developed, as depicted in Fig. 2.

The robotic structure is composed of an instrument capable of translational movement along the linear rail using the actuator M3, facilitating the insertion and extraction of the instrument through the center of the circular rail. The circular rail enables movement of the carriage through actuator M2. Actuator M1 facilitates rotational motion around the horizontal axis. The combination of these last two motions defines a virtual sphere on which the instrument operates.

Furthermore, the spherical robot incorporates a counterweight system, designed to dynamically counterbalance the load of the instrument based on the gravity compensation concept, the impact of the balancing mechanism's scaling being directly proportional to the inclination angle, increasing the stiffness of the mechanical structure.

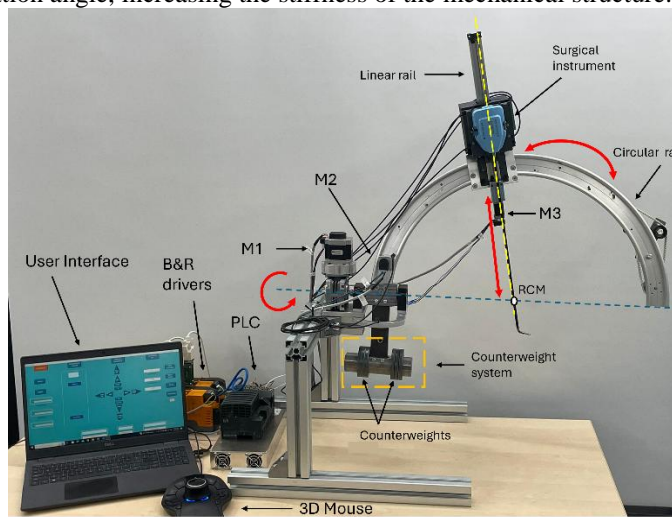

**Fig.2** Experimental model of the spherical robot

### 2.3    Control System Architecture

The architecture presented in Fig.3 is separated into two parts: one that controls the movements of the instrument such as pitch, yaw, roll, grasping, and pinching and another which assures the movement of the robotic system. This system features a modular design, enabling interchangeability and compatibility with a variety of MIS procedures, thus facilitating seamless integration.

**The User Interface Level** features a 3D Mouse Space for surgeon control, a PC for algorithm implementation and data analysis, and a Graphical User Interface offering control options, system status, and live stream from the endoscopic camera.

The **Command-and-Control Level** is composed of the control architecture for both the instrument/endoscopic camera attached to the arm and robotic structure. Commands are sent from the user interface through a TCP/IP connection to the Raspberry Pi Pico. From there, they are directed to the drivers responsible for coordinating movements. For the instrument/endoscopic camera attached to the spherical robot, the commands are translated into actions for the DC motors. Meanwhile, for the robotic structure, data is sent to the Programmable Logic Controller (PLC) via Modbus protocol. The PLC then communi-cates with the drivers, which in turn activates the stepper motors based on the received instructions according to the mathematical model of the robot.

This level incorporates essential components for instrument control, including a Raspberry Pi microcontroller, two MD35A drivers for actuators (each driver handling two actuators), a PLC, and two B&R drivers (one for NEMA 8 and NEMA 11 actuators, and another for NEMA 24).

The **Physical Level** includes four Polulu actuators for instrument movement, a NEMA 8 actuator facilitating linear carriage translation, a NEMA 11 actuator enabling translational movement on the circular rail, and a NEMA 24 actuator for rotational arm movement, ensuring the motion of the entire robotic structure. All motors are equipped with encoders. Additionally, it includes proximity sensors for homing and access limit detection, ensuring precise and controlled motion within the system.

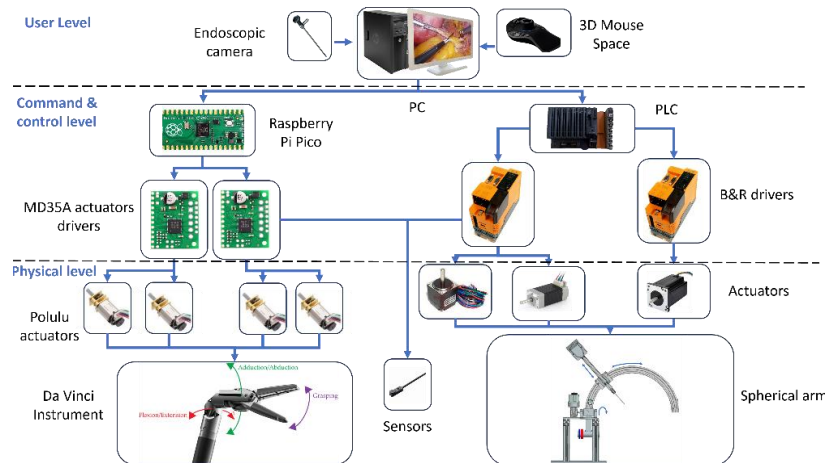

**Fig.3** Control architecture of the spherical robot

The control architecture operates on the master-slave concept, where the master console, serving as the primary control center, directs the actions of the slave console. In this configuration, the master console provides instructions and commands. The slave console, represented by the robotic structure, follows the commands received from the master console, executing precise movements and tasks accordingly.

## 3    Results

To functionally validate the experimental model a sequential motion was tested which involves the orientation of the endoscopic camera/instrument and then its insertion inside the patient.

Using same input data defined by current position of the endoscopic camera (A [243.1147mm, 172.3870mm, 105.8209 mm]) and the desired final position (B[263.6415 mmm, 57.6293 mm, -21.5508 mm])-Eq.(1)  two simulations using Siemens NX and two experimental runs were performed. The execution of a movement between the points A and B in the robot's workspace has been studied, as can be seen in Fig.4 The movement was executed in the first case by sequentially actuating the active joints (Fig.4-left), and in the second case by the simultaneous movement of active joints $q_1$ and $q_2$ (Fig.4-right), thus achieving the reorientation of the camera followed by its insertion. The results obtained during the NX simulation were processed and imported into MATLAB. During the experimental runs the position, velocity and acceleration of each active joint (motor) were recorded and the data was also imported in MATLAB. Data provided by the simulation using the sequential actuation of each active joint was graphically represented and the data coming from the experimental run was overlapped in Fig 5. Different colors were used to represent data from the simulation (red for positions, blue for velocity and green for accelerations) while for representing data from experimental runs a black dotted line was used. The same method was applied using the data extracted from simulation and experimental runs in the second case where the first two motors were actuated simultaneously.

The results of the overlapping between each two sets of data revealed no considerable differences between the experimental runs and the simulations, proofing that the motion of the robot is performed according to motion law implemented in the PLC, thus validating the functionality of the robot and of the control system.

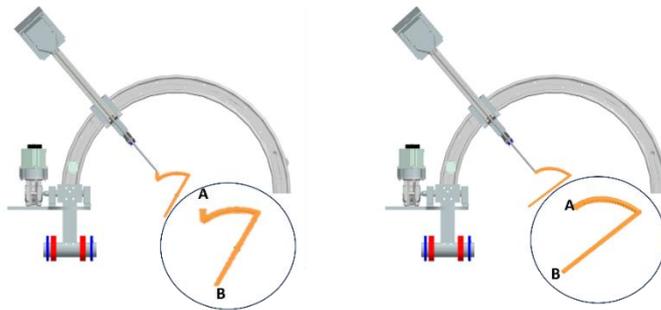

**Fig.4** Motion trajectories of the active joints (left – sequentially movement; right – simultaneously)

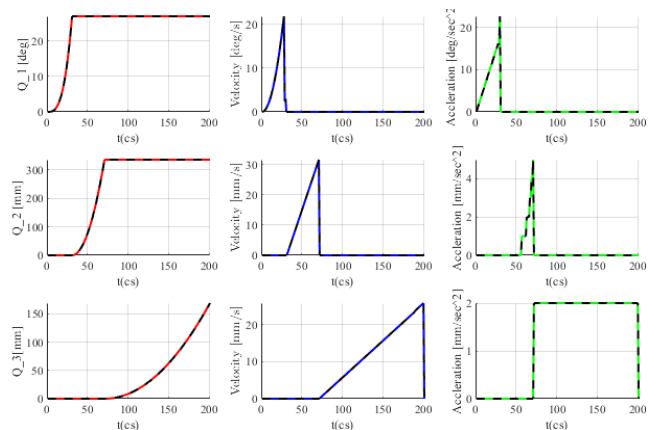

**Fig.5** Sequential motions executed by the active joints

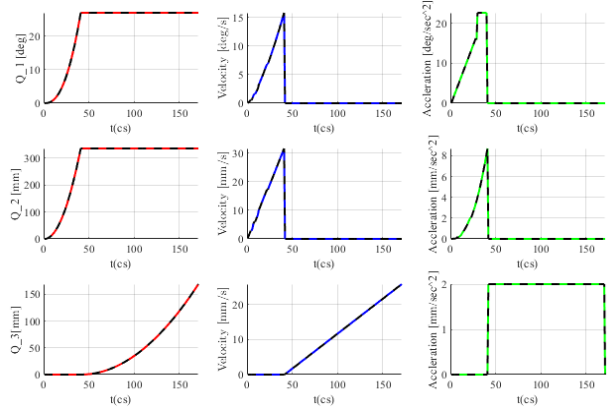

**Fig.6** Continuously motions executed by the active joints

## 4 Conclusions

The study presents an innovative spherical robot designed for MIS procedures, addressing key architectural constraints such as RCM compliance, system stiffness, and compact footprint. The control system, employing a master-slave architecture, has been successfully implemented and validated both through simulation in NX software and experimental testing. With its reduced dimension and stiffness, the proposed system holds promise for integration with other robotic systems, potentially enhancing overall surgical capabilities. Experimental validation involved sequential and simultaneous motion tests of an endoscopic camera/instrument's orientation and insertion, with two simulations in Siemens NX and two experimental runs. Data on position, velocity, and acceleration of active joints were recorded and compared between simulations and experiments, showing no significant differences and validating the robot's functionality and control system. Is important to notice that the RCM point is assured trough architectural constraint, thus the
Future work will focus on further exploration of this integration and its implications for advancing robotic-assisted surgery.

### Acknowledgment

This work was supported by the project New smart and adaptive robotics solutions for personalized minimally invasive surgery in cancer treatment - ATHENA, funded by European Union – NextGenerationEU and Romanian Government, under National Recovery and Resilience Plan for Romania, contract no. 760072/23.05.2023, code CF 116/15.11.2022, through the Romanian Ministry of Research, Innovation and Digitalization, within Component 9, investment I8 and by a grant of the Ministry of Research, Innovation and Digitization, CNCS/CCCDI—UEFISCDI, project number PN-III-P2-2.1-PED-2021-2790 694PED—Enhance, within PNCDI III.